\documentclass[conference]{IEEEtran}
\IEEEoverridecommandlockouts
\usepackage{cite}
\usepackage{amsmath,amssymb,amsfonts}
\usepackage{algorithmic}
\usepackage{textcomp}
\usepackage{xcolor}

\newcommand{\tabincell}[2]{\begin{tabular}{@{}#1@{}}#2\end{tabular}}
\usepackage[normalem]{ulem}
\usepackage{graphicx, amsthm, amsmath, amssymb}
\usepackage{subfig}
\usepackage[ruled,linesnumbered]{algorithm2e}
\SetAlCapNameFnt{\footnotesize}
\SetAlCapFnt{\footnotesize}
\usepackage{bm}
\usepackage{url}
\usepackage{footnote}
\usepackage{multirow}
\usepackage{longtable}

\def\BibTeX{{\rm B\kern-.05em{\sc i\kern-.025em b}\kern-.08em
    T\kern-.1667em\lower.7ex\hbox{E}\kern-.125emX}}
\DeclareRobustCommand*{\IEEEauthorrefmark}[1]{%
    \raisebox{0pt}[0pt][0pt]{\textsuperscript{\footnotesize\ensuremath{#1}}}}
\begin{document}

\title{PCNN: Pattern-based Fine-Grained Regular Pruning Towards Optimizing CNN Accelerators}
% \vspace{-20mm}

\author{\IEEEauthorblockN{Zhanhong Tan\IEEEauthorrefmark{1},
Jiebo Song\IEEEauthorrefmark{4}, Xiaolong Ma\IEEEauthorrefmark{2},
Sia-Huat Tan\IEEEauthorrefmark{1}, Hongyang Chen\IEEEauthorrefmark{3}, Yuanqing Miao\IEEEauthorrefmark{4}, Yifu Wu\IEEEauthorrefmark{4},\\ Shaokai Ye\IEEEauthorrefmark{4}, Yanzhi Wang\IEEEauthorrefmark{2}, Dehui Li\IEEEauthorrefmark{4}$^,$\footnotemark{$^*$}, Kaisheng Ma\IEEEauthorrefmark{1}$^,$\IEEEauthorrefmark{4}$^,$\footnotemark{$^*$}}
\IEEEauthorblockA{
% Department of Whatever,
% Whichever University\\
% Wherever\\
\IEEEauthorrefmark{1}Tsinghua University,
\IEEEauthorrefmark{2}Northeastern University,
\IEEEauthorrefmark{3}Xi'an Jiaotong University,\\
\IEEEauthorrefmark{4}Institute for Interdisciplinary Information Core Technology}\vspace{-0em}}

% Xiaolong Ma, Sia-Huat Tan, Hongyang Chen, Yuanqing Miao, Yifu Wu, Shaokai Ye, Yanzhi Wang, Dehui Li and Kaisheng Ma}

% \author{\IEEEauthorblockN{Zhanhong Tan}
% \IEEEauthorblockA{\textit{Institute for Interdisciplinary Information Sciences} \\
% \textit{Tsinghua University}\\
% Beijing, China \\
% tanzh19@mails.tsinghua.edu}
% \and
% \and
% \and
% \and

% }

\maketitle

\begin{abstract}
Weight pruning is a powerful technique to realize model compression.
We propose PCNN, a fine-grained regular 1D pruning method.
A novel index format called Sparsity Pattern Mask (SPM) is presented to encode the sparsity in PCNN.
Leveraging SPM with limited pruning patterns and non-zero sequences with equal length, PCNN can be efficiently employed in hardware.
Evaluated on VGG-16 and ResNet-18, our PCNN achieves the compression rate up to 8.4$\times$ with only 0.2\% accuracy loss.
We also implement a pattern-aware architecture in 55nm process, achieving up to 9.0$\times$ speedup and 28.39~TOPS/W efficiency with only 3.1\% on-chip memory overhead of indices.

\end{abstract}

% \begin{IEEEkeywords}
% Pattern learning , Pruning, Inference, Domain Specific Architecture
% \end{IEEEkeywords}

\section{Introduction}
Convolutional Neural Networks (CNN) have been developing rapidly in many applications, such as image classification~\cite{ImageNet}, object detection~\cite{object_detection} and natural language processing~\cite{nlp}.
When neural networks achieve higher accuracy with increasing computation and parameters~\cite{ResNet}~\cite{VGG}, many accelerators~\cite{DianNao}~\cite{Eyeriss}~\cite{UNPU}~\cite{FPGA2} implemented on ASIC or FPGA with abundant parallel units are proposed to deploy neural networks on edge devices.
Although these accelerators can improve throughput, it is still very challenging to deal with tremendous computation and transfer large amounts of data from DRAM to the on-chip memory.
%~\cite{DianNao,Eyeriss,UNPU,SCNN,Thinker,Cambricon-S,Winograd,FPGA1,FPGA2}
Weight pruning is an effective approach to reduce model size.
% ~\cite{Deep_Compression}~\cite{Prune0}.
%~\cite{Prune0}~\cite{Deep_Compression}.
%~\cite{prune_regular}~\cite{SSL}.
However, the methods proposed in early works~\cite{Deep_Compression}~\cite{Prune0} prune weights randomly (named irregular pruning). Irregular pruning needs to store weights in Compressed Sparse Column (CSC) format~\cite{EIE}, which leads to considerable extra indices to represent weights.
Besides, imbalance of workload among different computing units in the highly parallel architecture causes resource under-utilization, which prevents the hardware from fully leveraging the advantages of weight~pruning.

%\begin{figure}
%    \centering
%    \includegraphics[width=2.4in]{figures/pruning}
%    \caption{Illustration of channel pruning, filter pruning, kernel pruning and shape pruning.}
%    \label{figure:Different_Regular_Pruning}
%\end{figure}

Contrary to irregular pruning, regular pruning is more hardware-friendly.
Different granularities are explored in previous works, including irregular sparsity (0D), vector-level sparsity (1D), kernel-level sparsity (2D), and filter-level sparsity (3D)~\cite{prune_regular}~\cite{regular2}~\cite{SSL}.
However, the accuracy drops as the increment of pruning regularity~\cite{prune_regular}.
Therefore, it is imperative to find a new granularity to achieve regular pruning with high accuracy.

In this paper, we firstly propose Sparsity Pattern Mask (SPM), a novel kernel-level index format to indicate non-zero weights in each kernel.
Based on SPM, we present PCNN, a fine-grained regular 1-D pruning method with the identical number of non-zero weights in each kernel of one layer.
In this case, the computation workload of different convolution windows can be balanced with a few number of kernel patterns in each layer.
To push PCNN to be more regular, we further leverage multiple knapsack framework to distillate patterns (i.e., fewer patterns).
In this way, we can employ fewer bits to encode SPM. 

Based on PCNN, we implement a pattern-based architecture in 55nm process. Specialized memory optimization is proposed to map the PCNN-based workload with SPM code in hardware. Leveraging the benefit of PCNN, a sparsity-aware PE array is designed to achieve highly parallel computation with a delicate pipeline. Moreover, the sparsity-aware PE array can process sparse weights and activations simultaneously. 
% This hardware-algorithm co-design can give full play to the advantage of pruning, obtaining up to  

% In experiments, we explore the relationship between accuracy and various factors, such as sparsity and the number of patterns.
% Additionally, 
% This delicate design fully achieves distinct memory and computational reduction due to balanced workload in different convolution windows and fewer indices.

%In summary, we make the following contributions:
%\begin{itemize}
%    \item
   % We first propose Sparsity Pattern Mask (SPM), a novel kernel-level index format to encode non-zero weights in each kernel.
   % \item
  %  Based on SPM, we further propose Pattern-based Neural Network (PCNN), a fine-grained regular pruning method that is orthogonal to existing pruning approaches.
   % The proposed PCNN method is friendly to both accuracy and the hardware.
  %  We also formulate a multiple knapsack problem based framework to achieve pattern distillation which can reduce the number of patterns and obtain a more regular model. 
    %\item
  %  Taking advantage of PCNN, we present a pattern-aware architecture implemented in 55nm process with tapeout.
 %   A specialized memory optimization method is proposed to map PCNN with SPM in hardware.
    %In our architecture, the PE array can process sparsities of activations and weights efficiently with the help of PCNN and dataflow~design.
   % \item
   % Experiments show that the PCNN algorithm can achieve up to 8.65$\times$ compression ratio with only 0.2\% accuracy loss.
    In experiments, combined with other pruning methods like channel pruning, the PCNN algorithm can achieve up to 34.4$\times$ reduction of model size with negligible accuracy loss, which proves its orthogonality feature.
    With the power of PCNN, the proposed pattern-aware architecture fully leverages benefits from PCNN and shows up to 9$\times$ speedup and 28.39 TOPS/W efficiency with only 3.1\% memory overhead of indices.
%\end{itemize}
$\footnote{Corresponding authors are Dehui Li and Kaisheng Ma, and their emails are lidehui@iiisct.com and kaisheng@mail.tsinghua.edu.cn respectively.}$

\section{The Proposed PCNN Framework} \label{section:principle}

\subsection{Descriptions of PCNN}

\begin{figure}[!t]
\centering
\includegraphics[width=2.9in]{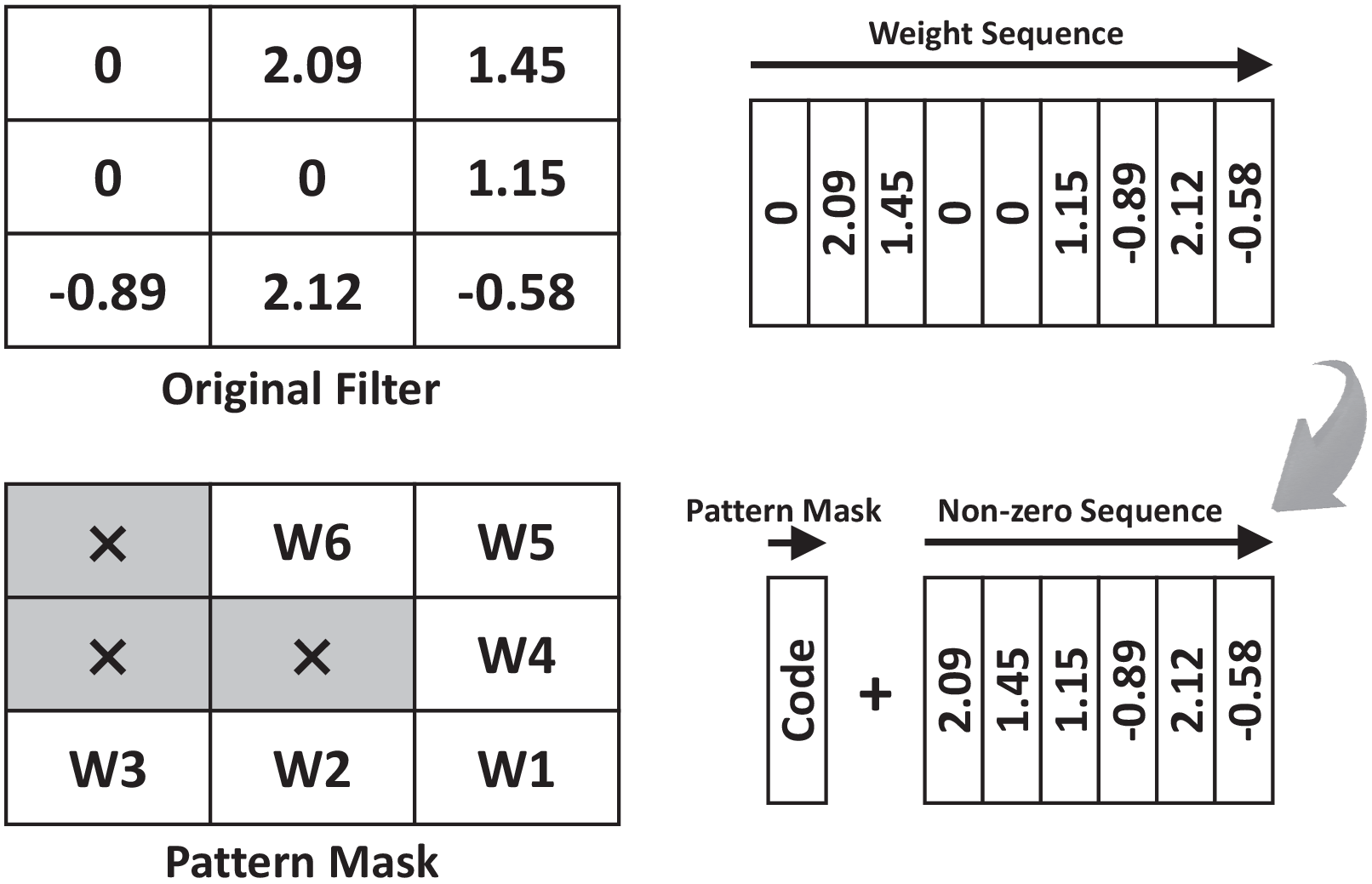}
\caption{Top: the original representation for a kernel. Bottom: pattern-based representation using an SPM index and the non-zero sequence.}
%\includegraphics[width=2.7in]{figures/encode.eps}
%\subfloat[The overhead and compression of various number of pattern types. The number of non-zero is 4.]
%{\includegraphics[width = 2.7in]{figures/typenum.eps}\label{figure:pattern-overhead}}
%\caption{Basic concept for pattern-filter and the result for various total of pattern types.}
\label{figure:pattern-explain}
\vspace{-1mm}
\end{figure}

Generally, there are some zeros in one convolution kernel.
To avoid storing zeros in the memory, we propose a kernel-level index format called Sparsity Pattern Mask (SPM) to encode sparse weights.
As shown in Figure~\ref{figure:pattern-explain}, the non-zero weights in a kernel are distributed in a specific pattern, which can be encoded with an SPM index.
As a result, only the non-zero values and the SPM index need to be stored.
Different from Compressed Sparse Column (CSC) format~\cite{EIE} which employs one index to each weight, we only need to apply one SPM index to each kernel.
In contrast, the index overhead is much smaller.

\begin{figure}[t]
    \centering
    \includegraphics[width=3.3in]{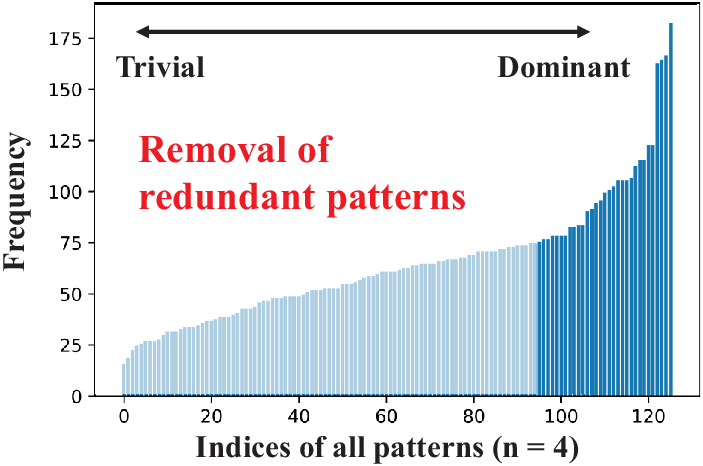}
    \caption{Pattern distribution in CONV4 of VGG-16, where the number of non-zero is 4. Thus there are 126 patterns in total.}
    \label{figure:distribution}
    \vspace{-2mm}
\end{figure}

However, there are $\sum_{i=0}^{9} \dbinom{9}{i}=512$ in total patterns in $3\times3$ kernels, indicating that the bitwidth of SPM index is 9.
Therefore, in order to reduce the bitwidth overhead of SPM encoding and simultaneously maintain the balanced workload for parallel computation, we propose PCNN which keeps identical sparsity in each kernel of one layer (i.e., the numbers of zeros in different kernels are the same).
In this way, the number of patterns is reduced to $\max \{ \dbinom{9}{0}, \dbinom{9}{1}, ..., \dbinom{9}{9} \}=126$.
In Figure~\ref{figure:distribution}, there are even some redundant patterns when we apply PCNN, which means that the number of patterns can be ulteriorly reduced to some~extent.

Based on the above considerations, we expect to find the optimal pruning manner to deploy PCNN with appropriate sparsity and the number of patterns.
Consequently, we establish a framework to describe PCNN with the following terminologies.
The set $S = \{ s_{1}, s_{2}, ..., s_{l} \}$ is the kernel sparsity of each convolution layer, determined by ${n}_{l}/{K}_{l}$, where ${n}_{l}$ denotes the number of non-zero values while ${K}_{l}$ is the kernel size for each layer. 

Next, the set $\bm{F}=\{ F_1, F_2, ..., F_l \}$ is a full collection of pattern sets for each layer and the extracted one $\bm{P}=\{ P_1,  P_2, ...,  P_l \}$ is extracted from $\bm{F}$ without redundant patterns as shown in Figure~\ref{figure:distribution}. 
Thus, in the PCNN learning framework, we intend to explore the appropriate $s_l$ and $P_l$ for each layer, aiming to achieve a smaller model with fewer~patterns.

\subsection{KP-based Pattern Distillation}
% ~\\
\textbf{Problem modeling}. As mentioned above, we intend to further reduce the number of patterns in our PCNN framework. Thus, we employ pattern distillation to choose the dominant ones. Pattern distillation means we have to select finite valuable patterns from a candidate set, which perfectly corresponds with the core idea of the knapsack problem. 

With a given Convolutional Neural Network containing $N$ convolution layers, $W_l$ denotes the weights in the $l$-th convolution layer and the set $\bm{W}=\{W_1,...,W_l,...,W_N\}$ is a weight collection of $N$ layers. $n$ denotes the number of non-zero weights in each kernel. We formulate pattern distillation~as:
% \begin{gather} \small
% \underset{x_{li}} {min} \sum_l \sum_j \left \| w_{lj} - \Pi_{P_l} ^{w_{lj}}\right \|_2, \\ 
%   s.t. \quad  \sum x_{li} \leqslant V_l, x_{li} \in \{0,1\}, i=1,...,t_n\\
% P_l = \bigcup p_{li} x_{li}, ~ p_{li} \in F_n,\\
% l=1,..., N, ~ j = 1,...,N_l,
% \end{gather}
\begin{align} \small \nonumber
&  \underset{x_{li}} {min} \sum_l \sum_j \left \| w_{lj} - \Pi_{P_l} ^{w_{lj}}\right \|_2, \\ 
&  s.t. \quad  \sum x_{li} \leqslant V_l, x_{li} \in \{0,1\}, i=1,...,t_n, \\ \nonumber
&  P_l = \bigcup p_{li} x_{li}, ~ p_{li} \in F_n,\\ \nonumber
&  l=1,..., N, ~ j = 1,...,N_l, \nonumber
\vspace{-8mm}
\end{align}
% \begin{equation} \small
% \centering
%     \begin{split}
%         \underset{x_{li}} {min} \sum_l \sum_j \left \| w_{lj} - \Pi_{P_l} ^{w_{lj}}\right \|_2, \\
%         s.t. \quad  \sum x_{li} \leqslant V_l, x_{li} \in \{0,1\}, i=1,...,t_n\\
%         P_l = \bigcup p_{li} x_{li}, ~ p_{li} \in F_n,\\
%         l=1,..., N, ~ j = 1,...,N_l,
%     \end{split}
% \end{equation}
% \begin{gather} 
% \underset{x_{li}} {min} \sum_l \sum_j \left \| w_{lj} - \Pi_{P_l} ^{w_{lj}}\right \|_2, \\ 
%   s.t. \quad  \sum x_{lj} \leqslant V_l, x_{lj} \in \{0,1\},\\
% P_l = \bigcup p_{li} x_{li}, ~ p_{li} \in F_n,\\
% l=1,..., N, ~ j = 1,...,N_l.
% \end{gather}

where $w_{lj}$ is the $j$-th kernel in $W_l$ and $N_l$ is the number of kernels in $W_l$. 
$F_n$ denotes the full set or candidate set of the patterns that have uniform sparsity and $t_n$ denotes the total number of patters in $F_n$ (i.e., $t_n = |F_n|$).
$P_l$ is the selected patterns in the $l$-th layer that is derived form $F_n$. Hyper-parameter $V_l$ is the desired number of the selected patterns ($V_l=| P_l|$).
The $i$-th pattern in $F_n$ to $P_l$ is selected when $x_{lj}=1$, and vice versa.
$\Pi^{w_{lj}}_{P_l}$ is a projection function that matches $w_{lj}$ to the nearest pattern in $P_l$ by keeping top $n$ absolute values.

The pattern distillation problem is similar to the knapsack problem (KP).
If we regard each $w_{lj}$ as a single knapsack, then the capacity of each knapsack is 1. 
In other words, we can only choose one pattern from the candidate set for $w_{lj}$.
Representing $\bm{W}$ with the selected patterns, the problem can be regarded as the multiple knapsack problem (MKP).
Particularly, since all capacities of $w_{lj}$ are 1, KP-based pattern selection is a multiple knapsack problem with identical capacities (MKP-1).

\textbf{Solution with greedy algorithm}.
In the case of our special problem, we propose an efficient greedy algorithm to solve KB-based pattern optimization.
For each layer, we first select the most valuable pattern for each $w_{lj}$ and then collect the number of patterns that have been chosen.
Finally, we keep the patterns that have the highest frequency (top $V_l$ patterns).
Details are shown in Algorithm~\ref{algorithm1}.
As a result, the set $P_l$ will contain a fewer number of patterns.
\begin{algorithm}[htp]\footnotesize
\textbf{Input:} $F_n$, $V_l$, $l=1,...,N$;\\
 \textbf{Initialization:} $P_l = \emptyset$, $l=1,...,N $; \\
 \For{$l \leftarrow 1$ \KwTo $N$}{
 $N_i=1, i=1,...,t_n;$\\
  \For{$j \leftarrow 1$ \KwTo $N_l$}{
\For{$i \leftarrow 1$ \KwTo $t_n$}{
	\If{$ w_{lj}$ is nearest to $p_i$}{
      		$N_i = N_i + 1$;
   	}	
}
}

$\{idx_1, ..., idx_{V_l}\} = max\_idx\_of\_sorted(N_1,...,N_{t_n})$;\\
\For{$i \leftarrow 1$ \KwTo $V_l$}{
 	$P_l = P_l \bigcup p_{idx_i}$;
 }
}
\textbf{Return:} $P_l$, $l=1,...,N$.
\caption{Pattern distillation algorithm}
\label{algorithm1}
\end{algorithm}
\section{The architecture for PCNN-Based Computing}

%In this section, we propose an architecture for PCNN. With the benefit of PCNN, the smaller size of memory is needed and high parallel computation can be achieved, giving full play of our model compression method.

\subsection{Mapping PCNN in Memory with SPM}

\begin{figure}[!htp]
\vspace{-3mm}
\centering
\subfloat[Overall architecture.]
 {\includegraphics[width=2.0in]{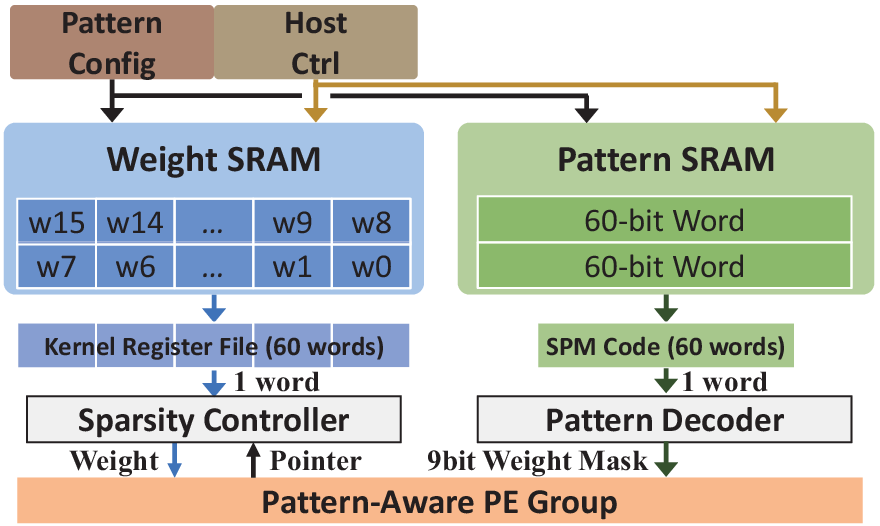}\label{figure:Overall}}
%  \hspace{0.03in}
\subfloat[Storing format for weights.]
 {\includegraphics[width=1.45in]{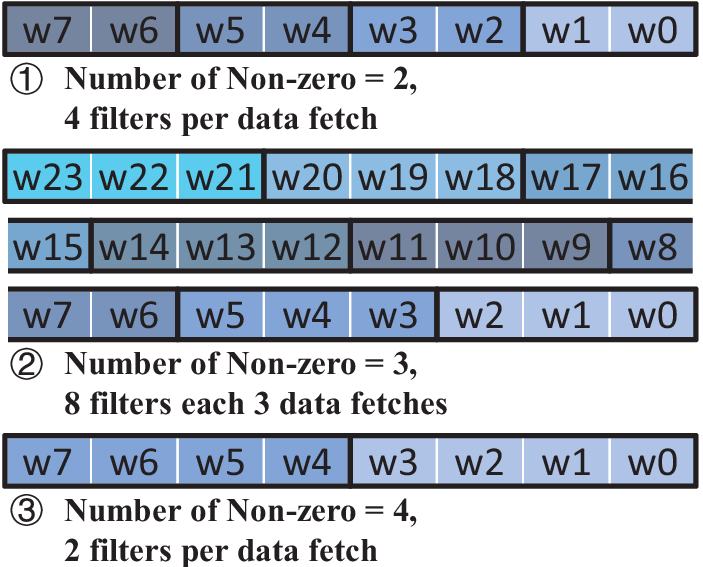}\label{figure:format}}
\caption{Memory design for weights and patterns in the pattern-aware architecture.}
\label{figure:mem_design}
\vspace{-5mm}
\end{figure}

Figure~\ref{figure:Overall} is the overall architecture for PCNN-based computation. Pattern Config (PaC) provides information of kernel sparsity and employs SPM mapping table for the decoder. 
In the pattern pruning framework, a kernel is denoted by a corresponding SPM code and a non-zero sequence, stored separately in Weight SRAM and Pattern SRAM.
For a $3\times3$ kernel, the length of the non-zero sequence ranges from 1 to 9.
Therefore, the sizes of kernel and SPM registers are 60-word which can integrally store kernels that contain 1 to 6 non-zero weights.
For other sparsities, we pad zeros to align the memory.
A kernel with non-zero weights is fetched into the register and the corresponding SPM code is simultaneously decoded into a 9-bit weight mask.
After generating weight pointers based on the mask, the pattern-aware PE group will properly fetch weights from the kernel register with sparsity pointers which will be presented in the next subsection.

Figure~\ref{figure:mem_design} is the memory layout for PCNN. In different sparsity cases, weights are similarly stored in order. The host controller can delicately access memory, fetching data to fill in the registers according to the kernel sparsity configured in~PaC.

% However, the word-length of pattern masks is different from that of weights and variable across layers. By contrast, we employ group division in bit-wise rather than sequence-wise in mask memory access. According to the result of section ~\ref{section:experiment}, actually at most 32 types of pattern are sufficient to guarantee the accuracy, so we store pattern codes in SRAM of 60 bit-width, as shown in Fig.~\ref{figure:format}, which includes 12 patterns of 5-bit (32 types), or 15 patterns of 4-bit (16 types), or 20 patterns of 3-bit (8 types), or 30 patterns of 2-bit (4 types), or 60 patterns of 1-bit (2 types). Pattern configurator here defines the pattern-word splitting manner to load the later shift register. Next, pattern code is sequentially read out to feed the PE array. Compared to fixed-length counterpart, the overhead of pattern pruning can be reduced to less than $\gamma \sum K_{l}$ as in section ~\ref{section:principle}. With dynamic bit-width format for pattern codes, the pattern overhead ratio in $l-$th layer is optimized from $\gamma$ to $\gamma_{l}^{'}$, where $\gamma_{l}^{'}$ is optimized to dynamic adjustment $\lceil{log_{2} V_{l}}\rceil$ rather than maximum method $\max \{ \lceil{log_{2} V_{1}}\rceil, \lceil{log_{2} V_{2}}\rceil, ..., \lceil{log_{2} V_{l}}\rceil\}$. The total overhead of pattern hence is reduced to $\sum \lceil{log_{2} V_{l}}\rceil K_{l}$. For example, in case of various setting $a$ on VGG-16 on CIFAR10 mentioned in section ~\ref{section:experiment}, dynamic scheme can save 40\% index storing compared to the maximum method.

\subsection{Sparsity-Aware Processing Element Group}

% \begin{figure}[t]
%     \centering
%     \includegraphics[width=3in]{figures/conv.png}
%     \caption{Computation of a convolution layer}
%     \label{figure:conv_computing}
% \end{figure}

\begin{figure}[!ht]
\centering
\subfloat[The details of datapath.]
{\includegraphics[width=1.67in]{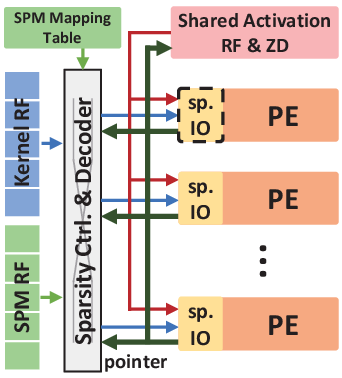}\label{figure:Computing-overview}}\hspace{0.05in}
\subfloat[Sparsity pointer generation.]
{\includegraphics[width = 1.67in]{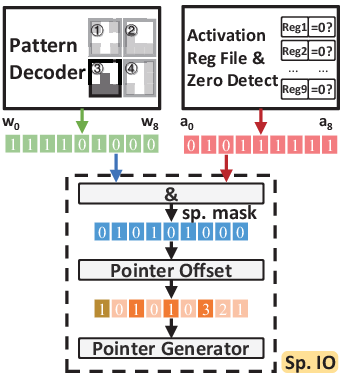}\label{figure:mask-generation}}\\
\subfloat[The calculation of pointer offset.]
{\includegraphics[width = 3.35in]{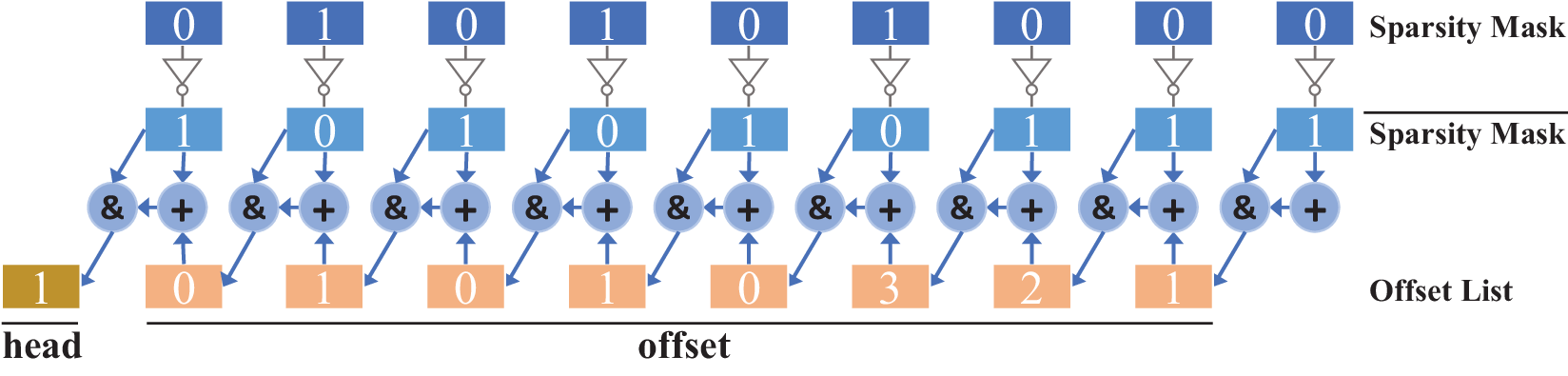}\label{figure:offset}}
\caption{Overview of the pattern-aware PE group.}
\label{figure:PE}
\vspace{-4mm}
\end{figure}

\begin{figure}[!ht]
    \centering
    \includegraphics[width=3.35in]{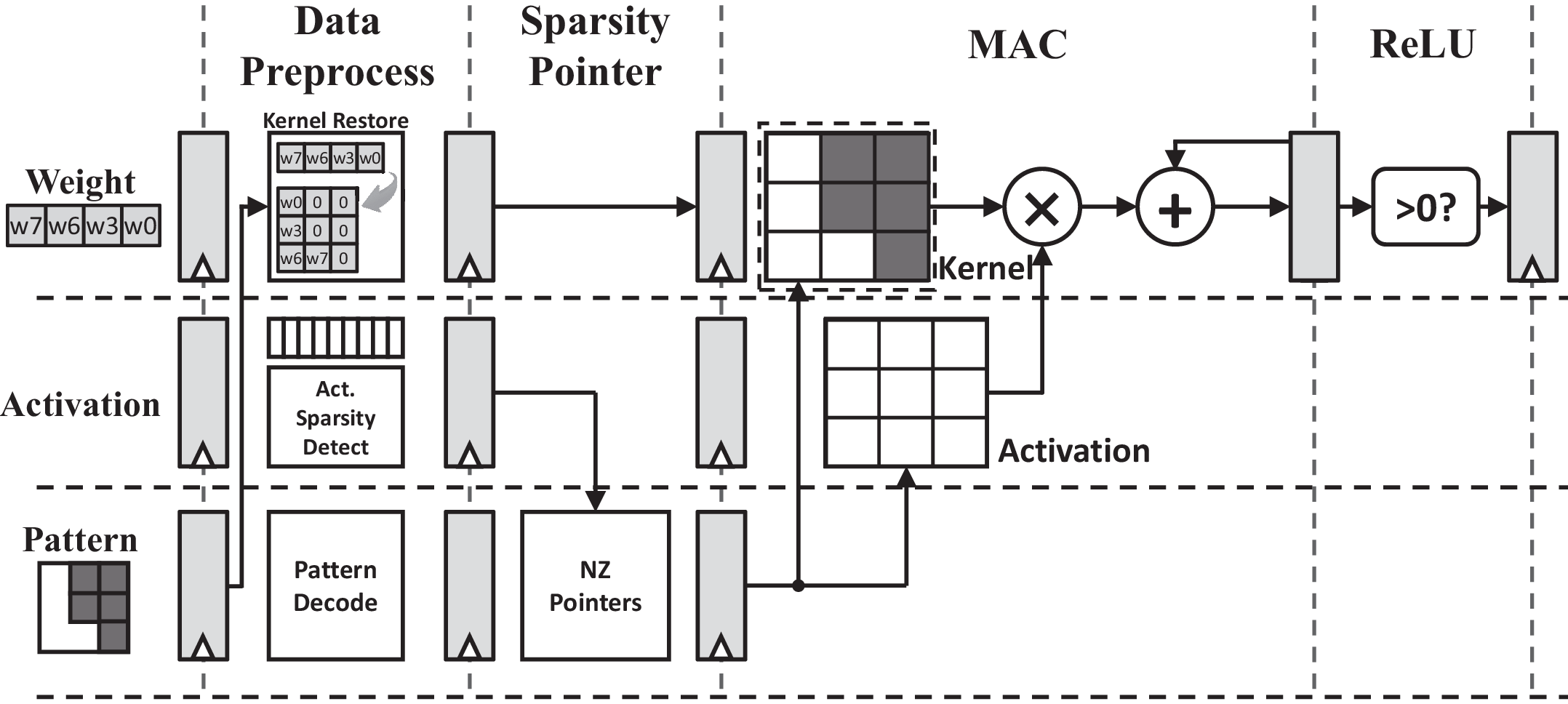}
    \caption{Pipeline strategy for pattern-aware architecture.}
    \label{figure:Pipeline}
    \vspace{-6mm}
\end{figure}

A detailed overview of the pattern-aware PE group is shown in Figure~\ref{figure:PE}, where the sparsity IO can generate pointers based on the weight mask from the SPM decoder to fetch weights properly. In our design, we implement 64 PEs with 4 MAC units in each one. Consequently, our architecture can perform at most 256 MACs per cycle.
Besides weight sparsity, we also leverage activation sparsity to further improve computing efficiency. Therefore, we employ shared-activation datapath to balance the workload from activations.

The sparsity pointer generation is shown in Figure~\ref{figure:mask-generation}. Both the weight mask and the activation mask are transferred to the sparsity IO and then the sparsity mask is generated. Later, pointer offsets can be obtained with the sparsity mask (Figure~\ref{figure:offset}).
There is an adder--AND chain to attain nine offsets, each of which denotes the distance to the nearest zero. The offsets can be elaborated as follows. Firstly, NOT operation is applied to each bit of sparsity mask. Secondly, we can obtain the pointer offsets by accumulating the number of zeros between every two non-zero weights.
With pointer offsets, we can attain the corresponding pointers to fetch the needed weights from the kernel register.
With the help of PCNN and the shared-activation dataflow, the workloads of weights and activations in different PEs are identical, contributing to higher resource utilization and better parallelism. 
% Certainly, the final workload depends on the sparsity mask so that occasionally there is an imbalance~case.

Figure~\ref{figure:Pipeline} shows the pipeline strategy in the proposed pattern-aware PE. In order to achieve high throughput, all the operations are pipelined. The first stage is the data pre-process stage. Weights are restored to the original kernel according to the SPM indices. Activations in a convolution window are loaded into the registers and activation sparsity masks are generated simultaneously. In the second stage, non-zero pointers are generated with the calculated offsets, which will select the effectual workload in the next stage to perform MACs. Last, when all partial sums from various input channels are added up together, ReLU is employed to attain the final result.
\begin{table*}[!t]
\centering
\setlength{\tabcolsep}{0.9mm}
\begin{tabular}{llllllll}
\hline
\multicolumn{1}{c|}{\textbf{Benchmark}} & \multicolumn{1}{c}{\textbf{Top1 acc}} & \multicolumn{1}{c}{\textbf{Top1 acc Loss}} & \multicolumn{1}{c}{\textbf{CONV FLOPs}} & \multicolumn{1}{c}{\textbf{FLOPs Pruned}} & \multicolumn{1}{c}{\textbf{CONV Parameters}} & \multicolumn{1}{c}{\textbf{\begin{tabular}[c]{@{}c@{}}Compression\\ (weight)\end{tabular}}} & \multicolumn{1}{c}{\textbf{\begin{tabular}[c]{@{}c@{}}Compression\\ (weight+idx)\end{tabular}}} \\ \hline
\hline
\multicolumn{1}{l|}{\textbf{VGG-16, Baseline}} & 93.54\% & - & 3.13$\times10^8$ & - & 1.47$\times10^7$ & - & - \\ \hline
\multicolumn{1}{l|}{\textbf{VGG-16, n = 4}} & 93.79\% & +0.25\% & 1.39$\times10^8$ & 56.5\% & 0.65$\times10^7$ & 2.3$\times$ & 2.2$\times$ \\ \hline
\multicolumn{1}{l|}{\textbf{VGG-16, n = 3}} & 93.58\% & +0.04\% & 1.04$\times10^8$ & 66.7\% & 0.49$\times10^7$ & 3.0$\times$ & 2.9$\times$ \\ \hline
\multicolumn{1}{l|}{\textbf{VGG-16, n = 2}} & 93.52\% & -0.02\% & 0.30$\times10^8$ & 77.8\% & 0.33$\times10^7$ & 4.5$\times$ & 4.1$\times$ \\ \hline
\multicolumn{1}{l|}{\textbf{VGG-16, n = 1}} & 93.07\% & -0.21\% & 0.35$\times10^8$ & 88.9\% & 0.16$\times10^7$ & 9.0$\times$ & 8.4$\times$ \\ \hline
\multicolumn{1}{l|}{\textbf{VGG-16, Various setting$^a$}} & 93.33\% & -0.21\% & 0.35$\times10^8$ & 88.8\% & 0.16$\times10^7$ & 9.0$\times$ & 8.4$\times$ \\ \hline
\end{tabular}
\\
\footnotesize{$^a$ n in various layers: 2-1-1-1-1-1-1-1-1-1-1-1-1 with 32 patterns in n = 2 layers and 8 patterns in n = 1 layers.}\\
  \caption{Pruning rate and accuracy of different n for VGG-16 on CIFAR10.}
  \label{table:ex-Nz-vg}
  \vspace{-3mm}
\end{table*}

\begin{table*}[!t]
\centering
\setlength{\tabcolsep}{0.9mm}
\begin{tabular}{llllllll}
\hline
\multicolumn{1}{c|}{\textbf{Benchmark}} & \multicolumn{1}{c}{\textbf{Top1 acc}} & \multicolumn{1}{c}{\textbf{Top1 acc Loss}} & \multicolumn{1}{c}{\textbf{CONV FLOPs}} & \multicolumn{1}{c}{\textbf{FLOPs Pruned}} & \multicolumn{1}{c}{\textbf{CONV Parameters}} & \multicolumn{1}{c}{\textbf{\begin{tabular}[c]{@{}c@{}}Compression\\ (weight)\end{tabular}}} & \multicolumn{1}{c}{\textbf{\begin{tabular}[c]{@{}c@{}}Compression\\ (weight+idx)\end{tabular}}} \\ \hline
\hline
\multicolumn{1}{l|}{\textbf{ResNet-18, Baseline}} & 96.58\% & - & 5.55$\times10^8$ & - & 1.12$\times10^7$ & - & - \\ \hline
\multicolumn{1}{l|}{\textbf{ResNet-18, n = 4}} & 96.58\% & +0.06\% & 2.50$\times10^8$ & 54.5\% & 0.51$\times10^7$ & 2.2$\times$ & 2.1$\times$ \\ \hline
\multicolumn{1}{l|}{\textbf{ResNet-18, n = 3}} & 96.38\% & -0.20\% & 1.89$\times10^8$ & 65.5\% & 0.38$\times10^7$ & 3.0$\times$ & 2.8$\times$ \\ \hline
\multicolumn{1}{l|}{\textbf{ResNet-18, n = 2}} & 96.15\% & -0.43\% & 1.28$\times10^8$ & 76.7\% & 0.26$\times10^7$ & 4.3$\times$ & 4.0$\times$ \\ \hline
\multicolumn{1}{l|}{\textbf{ResNet-18, n = 1}} & 95.55\% & -1.03\% & 0.67$\times10^8$ & 88.0\% & 0.14$\times10^7$ & 7.9$\times$ & 7.3$\times$ \\ \hline
\multicolumn{1}{l|}{\textbf{ResNet-18, Various setting$^a$}} & 95.83\% & -0.75\% & 0.86$\times10^8$ & 84.5\% & 0.14$\times10^7$ & 7.9$\times$ & 7.3$\times$ \\ \hline
\end{tabular}
\\
\footnotesize{$^a$ n in various layers in bottle: 2-2-2-1-1-1-1-1-1-1-1-1-1-1-1-1 with 32 patterns in n = 2 layers and 8 patterns in n = 1 layers.}\\
  \caption{Pruning rate and accuracy of different n for ResNet-18 on CIFAR10.}
  \label{table:ex-Nz-res}
  \vspace{-3mm}
\end{table*}

\begin{table*}[!t]
  \centering
  \setlength{\tabcolsep}{1.6mm}
\begin{tabular}{llllllll}
\hline
\multicolumn{1}{c|}{\textbf{Benchmark}} & \multicolumn{1}{c}{\textbf{Top1 acc}} & \multicolumn{1}{c}{\textbf{Top1 acc Loss}} & \multicolumn{1}{c}{\textbf{CONV FLOPs}} & \multicolumn{1}{c}{\textbf{FLOPs Pruned}} & \multicolumn{1}{c}{\textbf{CONV Parameters}} & \multicolumn{1}{c}{\textbf{\begin{tabular}[c]{@{}c@{}}Compression\\ (weight)\end{tabular}}} & \multicolumn{1}{c}{\textbf{\begin{tabular}[c]{@{}c@{}}Compression\\ (weight+idx)\end{tabular}}} \\ \hline
\hline
\multicolumn{1}{l|}{\textbf{VGG-16, Baseline}} & 92.10\% & - & 6.82$\times10^9$ & - & 1.47$\times10^7$ & - & - \\ \hline
\multicolumn{1}{l|}{\textbf{VGG-16, n = 5}} & 92.47\% & +0.37\% & 0.85$\times10^9$ & 44.4\% & 0.82$\times10^7$ & 1.8$\times$ & 1.7$\times$ \\ \hline
\multicolumn{1}{l|}{\textbf{VGG-16, n = 4}} & 92.45\% & +0.35\% & 0.68$\times10^9$ & 56.5\% & 0.65$\times10^7$ & 2.3$\times$ & 2.2$\times$ \\ \hline

\end{tabular}

  \caption{Pruning rate and accuracy of different n for VGG-16 on ImageNet.}
  \label{table:ex-Nz-imagenet}
  \vspace{-5mm}
\end{table*}

\section{Evaluation}
\label{section:experiment}

\subsection{Methodology}

\textbf{Setup for evaluating the proposed PCNN}.
We summarize our PCNN results for CNN model compression on a series of benchmarks, including VGG-16~\cite{VGG} on CIFAR10~\cite{CIFAR} and ImageNet~\cite{ImageNet}, and ResNet-18~\cite{ResNet} on CIFAR10.
We initialize our learning framework with pre-trained models on PyTorch, following the pattern distillation.
After that, an Alternating Direction Method of Multipliers (ADMM)~\cite{boyd2011distributed} is employed to fine-tune our model.
Considering that convolution layers are getting more and more dominant at present, we mainly focus on convolution layers. 

\textbf{Setup for evaluating our PCNN-based architecture}.
Based on PCNN encoded with SPM, we implement the pattern-aware architecture with RTL and evaluate VGG-16 based on PCNN with VCS to obtain the performance.
The area and power of our architecture are attained with Design Compiler in UMC 55nm standard power CMOS process.

\subsection{Evaluating the Kernel Sparsity and the Number of Patterns}

In this part, we study different choices of kernel sparsity for VGG-16 and ResNet-18.
In ResNet-18, we only process the layers with 3$\times$3 filters and ignore 1$\times$1 ones which are too accuracy-sensitive.
We set n as 1, 2, 3, and 4 in all layers with at most 8, 32, 32, and 32 patterns respectively.

Table~\ref{table:ex-Nz-vg} shows the pattern pruning results of various n on the VGG-16 model on CIFAR10. 
PCNN leads to less than 0.5\% accuracy loss even when there is one weight left in each filter.
When we apply a different sparsity setting over layers, accuracy can be improved with similar compression rates and~speedup.

Similar results can be achieved for ResNet-18 on CIFAR10 as shown in Table~\ref{table:ex-Nz-res}. 
Within 0.5\% accuracy loss, pattern pruning achieves the compression rate ranging from 2.1$\times$ to 4.0$\times$ with the unified sparsity setting (n = 4, 3, and 2). 
Also, when various sparsity settings are applied, we can achieve better performance than the unified counterpart with a similar compression rate of 9$\times$. 
Furthermore, the results for VGG-16 on ImageNet are shown in Table~\ref{table:ex-Nz-imagenet} and we achieve 1.8$\times$ compression rate and 2.3$\times$ speedup with no harm to~accuracy.

Note the last columns of Table~\ref{table:ex-Nz-vg}$\sim$\ref{table:ex-Nz-imagenet} containing the actual compression rate considering the overhead of indies.
With PCNN, there are marginal compression rate drops due to kernel-level SPM indices.
On the contrary, for irregular pruning, taking VGG-16 with n = 4 as an example, the actual compression rate is 2.0$\times$, three times as low as ours. 

Later, we further restrain the number of patterns in each layer to study how regularity impacts accuracy.
As Table~\ref{table:ex-pattern-comparison} shows, we evaluate n = 4 and n = 2 with full patterns, 32, 16, 8, and 4 patterns. We use full patterns as baselines with 93.79\% and 93.52\% for n = 4 and n = 2 respectively and the weight compression rates are 2.3$\times$ and 4.5$\times$. 
The results show that there are no obvious accuracy drops with fewer patterns, which can help us to save the overhead of index storing. While in the higher sparsity case, the accuracy is more sensitive to the decrease of patterns. Actually, in the cases with high sparsity, we do not need to focus too much on the number of patterns because the loss of compression is little with SPM in PCNN. Averagely, 16 patterns are enough to maintain accuracy with less index overhead.

\begin{table}[!b]
  \vspace{-5mm}
  \centering
  \setlength{\tabcolsep}{1.8mm}
  \begin{tabular}{clll}
  \hline
  \multicolumn{2}{c}{\textbf{Benchmark}}                          & \textbf{Relative acc} & \tabincell{c}{\textbf{Compression}\\\textbf{(weight+idx)}} \\%hongyang winner
  \hline
  \hline
  \multicolumn{1}{c|}{\multirow{5}{*}{\textbf{n = 4}}} & \tabincell{l}{\textbf{$|P_{n}|$}=126 (full)\\\textbf{baseline}} & -                   & 2.14$\times$                     \\ 
  \cline{2-4}
  \multicolumn{1}{c|}{}                              & \textbf{$\left|P_{n}\right|$} = 32 & +0.32\%                   & 2.18$\times$                      \\
  \cline{2-4}
  \multicolumn{1}{c|}{}                              & \textbf{$\left|P_{n}\right|$} = 16 & +0.10\%                   & 2.20$\times$                     \\
  \cline{2-4}
  \multicolumn{1}{c|}{}                              & \textbf{$\left|P_{n}\right|$} = 8 & -0.05\%                   & 2.21$\times$                      \\
  \cline{2-4} 
  \multicolumn{1}{c|}{}                              & \textbf{$\left|P_{n}\right|$} = 4 & -0.17\%                   & 2.23$\times$                      \\
  \hline
  \multicolumn{1}{c|}{\multirow{5}{*}{\textbf{n = 2}}} & \tabincell{l}{\textbf{$|P_{n}|$}=36 (full)\\\textbf{baseline}} & -                   & 4.08$\times$                 \\%shengfa winner
  \cline{2-4}
  \multicolumn{1}{c|}{}                              & \textbf{{$|P_{n}|$}}  = 32 & +0.00\%                   & 4.13$\times$                      \\
  \cline{2-4}
  \multicolumn{1}{c|}{}                              & \textbf{{$|P_{n}|$}}  = 16& -0.22\%                   & 4.19$\times$                      \\
  \cline{2-4}
  \multicolumn{1}{c|}{}                              & \textbf{{$|P_{n}|$}}  = 8 & -0.54\%                   & 4.26$\times$                      \\
  \cline{2-4}
  \multicolumn{1}{c|}{}                              & \textbf{{$|P_{n}|$}}  = 4 & -0.71\%                   & 4.32$\times$                      \\
  \hline
  \end{tabular}
  \caption{Comparison of $|P_{n}|$ for VGG-16 on CIFAR10. If no index, the compression is 2.3$\times$ and 4.5$\times$ for n=4 and n=5.}
  \label{table:ex-pattern-comparison}
  \vspace{-1mm}
\end{table}

% The reasons for the results are two folds. On one hand, the overall trend is reasonable that less patterns contribute to be more regular, which relatively harms accuracy as Mao \emph{et al.}~\cite{prune_regular} illustrate. On the other hand, the difference in performance between two cases is because there are totally 126 kinds of patterns in the case where n = 4, while there are only 36 kinds in the case where n = 2. When we constraint unified sparsity of filters in one layer, the total $C_{K}^{n}$ patterns are consequently introduced, among which there are some less informative patterns with worse representation. After performing pattern distillation to extract more informative patterns, accuracy can be boosted efficiently. By contrast, in the case where n = 2, for there are very limited patterns of 36, less redundancy exists in pattern types and tends to be more sensitive to pattern selection. Consequently, the case where n = 2 is vulnerable to pattern selection. But actually, there is no need to reduce too many patterns when n = 2, because we consider that averagely 4-bit cost for each filter is reasonable enough for hardware. Compared to Deep Compression~\cite{Deep_Compression} with 4bit for each weight, our method achieves less overhead for storing indices.

\subsection{Comparison to Other Regular Compression Methods}

In this part, we compare PCNN to other pruning methods in other works for VGG-16 and ResNet-18 on CIFAR10.
For various baselines used in different works, we employ the accuracy loss relative to the respective baseline.
The comparison for VGG-16 is shown in Table~\ref{table:ex-Nz-VGG}.
Our method can remarkably compress parameters and reduce FLOPs simultaneously with negligible accuracy loss.
As for ResNet-18 shown in Table~\ref{table:ex-Nz-res-com}, PCNN also achieves better performance than other coarse-grained pruning. Especially, PCNN enjoys higher FLOPs reduction.%句型读起来有问题

% The better achievements of computation FLOPs reduction derive from higher sparsity in layers with large size, which takes up more calculated quantities, while other coarse-grained methods avoid pruning these layers.

\begin{table}[!b]
  \centering
  \setlength{\tabcolsep}{1.1mm}
  \begin{tabular}{lllllll}
    \hline
    \textbf{Benchmark} & \textbf{Relative acc} & \textbf{FLOPs} & \textbf{Compression}\\
    \hline
    \hline
    % \textbf{PCNN, n = 4, $\left|P_{n}\right|$} = 32 & +0.62\% & 55.6\% & 2.25$\times$ & 55.6\% & 2.25$\times$\\
    % \hline
    % \textbf{PCNN, n = 3, $\left|P_{n}\right|$} = 84 & +0.01\% & 66.7\% & 3.00$\times$ & 66.7\% & 3.00$\times$\\
    % \hline
    % \textbf{PCNN, n = 2, $\left|P_{n}\right|$} = 32 & -0.24\% & 77.8\% & 4.50$\times$ & 77.8\% & 4.50$\times$\\
    % \hline
    % \textbf{PCNN, Various setting, $\left|P_{n}\right|$} = 32 & -0.20\% & 88.4\% & 8.65$\times$ & 84.0\% & 8.65$\times$\\
    % \hline
    \textbf{PCNN} & +0.04\% & 66.7\% & 3.0$\times$\\
    \hline
    \textbf{PCNN} & -0.21\% & 88.8\% & 9.0$\times$\\
    \hline
    % \hline
    \textbf{Filter pruning \cite{Depth-1}} & +0.15\% & 33.3\% & 2.8$\times$\\
    \hline
    \textbf{Network slimming \cite{Depth-2}} & +0.14\% & 51.0\% & 8.7$\times$\\
    \hline
    % \textbf{SSS \cite{Depth-3}} & 0.00\%  & 19.8\% & 1.3$\times$\\
    % \hline
    % \textbf{SSS \cite{Depth-3}} & -0.09\% & 28.6\% & 1.8$\times$ \\
    % \hline
    % \textbf{try-and-learn b=0.5 \cite{Depth-4}} & -0.60\% & 83.3\% & 1.8$\times$\\
    % \hline
    \textbf{try-and-learn b=1 \cite{Depth-4}} & -1.10\% & 82.7\% & 2.2$\times$\\
    \hline
    \textbf{IKR~\cite{IKR}$^a$} & -0.90\% & 84.7\% & 4.3$\times$\\
    \hline
  \end{tabular}
  \\
  \footnotesize{$^a$ A VGG-16 inspired CNN containing 6 convoution and 2 FC layers}\\
  \caption{Comparison of various regular compression methods for VGG-16 on CIFAR10.}
  \label{table:ex-Nz-VGG}
  \vspace{-6mm}
\end{table}

\begin{table}[!ht]
  \centering
  \setlength{\tabcolsep}{1.4mm}
  \begin{tabular}{lllllll}
    \hline
    \textbf{Benchmark} & \textbf{Relative acc} & \textbf{FLOPs} & \textbf{Compression}\\
    \hline
    \hline
    \textbf{PCNN} & -0.20\% & 65.5\% & 3.0$\times$\\
    \hline
    \textbf{PCNN} & -0.75\% & 84.5\% & 7.9$\times$\\
    \hline
    % \hline
    % \textbf{Band-limited \cite{resnet-2}} & -0.27\% & - & 1.1$\times$\\
    % \hline
    \textbf{Band-limited \cite{resnet-2}} & -1.67\% & - & 2.0$\times$\\
    \hline
    % \textbf{try-and-learn b=2 \cite{Depth-4}} & -1.70\% & 64.7\% & 2.2$\times$\\
    % \hline
    \textbf{try-and-learn b=4 \cite{Depth-4}} & -2.90\% & 76.0\% & 4.6$\times$\\
    \hline
  \end{tabular}
  \caption{Comparison of various regular compression methods for ResNet-18 on CIFAR10.}
  \label{table:ex-Nz-res-com}
\end{table}

\subsection{The Orthogonality to Other Compression Methods}

\begin{table}[t]
  \centering
  \setlength{\tabcolsep}{2mm}
  \begin{tabular}{lllllll}
    \hline
    \textbf{Benchmark} & \textbf{Relative acc} & \textbf{Compression}\\
    \hline
    \hline
    \textbf{PCNN n = 5} & +0.38\% & 1.8$\times$\\
    \hline
    \textbf{\tabincell{l}{PCNN n = 5\\ + Kernel Pruning-A}} & +0.28\% & 4.4$\times$\\
    \hline
    \textbf{\tabincell{l}{PCNN n = 5\\ + Kernel Pruning-B}} & -0.27\% & 7.3$\times$\\
    \hline
  \end{tabular}
  \caption{Combined with kernel-level pruning for VGG-16 on ImageNet.}
  \label{table:ex-Nz-imagenet-kernel}
  \vspace{-2mm}
\end{table}

\begin{table}[t!]
  \centering
  \setlength{\tabcolsep}{2.1mm}
  \begin{tabular}{lllllll}
    \hline
    \textbf{Benchmark} & \textbf{Relative acc} & \textbf{Compression}\\
    \hline
    \hline
    \textbf{\tabincell{l}{PCNN + Channel\\Pruning-A}} & -0.02\% & 34.4$\times$\\
    \hline
    \textbf{\tabincell{l}{PCNN + Channel\\Pruning-B}} & -0.46\% & 50.3$\times$\\
    \hline
    % \textbf{Channel Pruning} & -1.14\% & 34.3$\times$\\
    % \hline
    \textbf{Structured ADMM~\cite{StructADMM}} & -0.60\% & 50.0$\times$\\
    \hline
    \textbf{SNIP~\cite{VGG-channel}} & -0.45\% & 20.0$\times$\\
    \hline
    \textbf{Synaptic Strength~\cite{Synaptic-Strength}} & +0.43\% & 25.0$\times$\\
    \hline
  \end{tabular}
  \caption{Combined with channel-level pruning for VGG-16 on CIFAR10.}
  \label{table:ex-Nz-cifar-filter}
  \vspace{-5mm}
\end{table}

% The proposed PCNN method can be combined with other pruning methods to achieve higher compression, where PCNN serves as a fine portion of pruning to maintain accuracy.

\textbf{Combined with kernel level pruning.} As shown in Table~\ref{table:ex-Nz-imagenet-kernel}, we perform some experiments for VGG-16 on ImageNet in the case where n = 5. We apply 2.4$\times$ and 4.1$\times$ kernel pruning with 1.8$\times$ compression rate from PCNN to achieve fused pruning.
Results show that the PCNN is orthogonal to kernel pruning and the combination of them achieves promising results.

\textbf{Combined with channel level pruning.}
Channel level pruning has been proved to be more regular with a higher compression rate but less friendly to accuracy~\cite{prune_regular}.
As results shown in Table~\ref{table:ex-Nz-cifar-filter}, the fused compression achieves 34.4$\times$ compression rate contributed by 3.75$\times$ PCNN compression and 9$\times$ channel pruning.
Therefore, PCNN is also orthogonal to channel pruning.
% In this case, channel pruning serves as a coarse-grained method to obtain higher sparsity while pattern pruning serves as a fine-grained pruning method to maintain the accuracy. Note that we use the same coarse-grained pruning method as structured ADMM~\cite{StructADMM}, and better result is obtained with pattern pruning. 

% It is observed that single method of coarse pruning does easily induce accuracy drop in high compression ratio. Compared to other works, the fused pruning method outperforms in accuracy and compression. 
% Consequently, pattern pruning is a promising pruning method to help coarse-grained pruning to achieve better accuracy.

\subsection{Evaluation on Pattern-Aware Architecture}
% We implement the pattern-aware architecture and the baseline counterpart with RTL. The area and power overhead of the pattern-aware architecture are evaluated with Synopsys Design Compiler in UMC 55nm standard power CMOS process.

% The main advantage of our pattern-aware architecture is summarized as follows. Firstly, it can deal with finer-grained pruning compared to channel pruning with extra 6.9\% memory costs, 15.8\% area and 17.8\% power overhead. Secondly, this architecture supports both activation sparsity and weight sparsity. Thirdly, it can be scaled for dense networks in the unified framework via pipeline bypassing and power gating.

\begin{figure}[!t]
    \centering
    \includegraphics[width=1.8in]{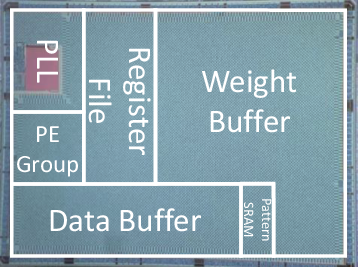}
    \caption{The layout of pattern-aware architecture.}
    \label{figure:layout}
    \vspace{-6mm}
\end{figure}

{\bf The overhead evaluation.}
The layout of our pattern-aware architecture is shown in Figure~\ref{figure:layout} and the overhead of each component is listed in Table~\ref{table:layout}. The power consumption is measured under the circumstance of 300~MHz on-chip frequency and 1~V voltage.
Contrary to irregular pruning, PCNN requires an SPM index for each kernel rather than weight.
With PCNN, a 128KB weight SRAM is employed holding up to 32768 kernels of 3$\times$3 size with 4 non-zeros with 8-bit quantization for common cases.
In this case, a 4K pattern SRAM is enough for the workload with 16 patterns in each layer.
As shown in analyses in section~\ref{section:experiment}, 16 patterns are sufficient to maintain accuracy.
Consequently, this architecture introduces only 3.1\% memory overhead to store indices.
The pattern SRAM only takes up 2.4\% area and 1.9\% power of the whole chip.
according to Table~\ref{table:layout}.
In comparison to the irregular pruning based architecture like EIE~\cite{EIE}, 64KB index SRAM is needed to denote 128K weights.

{\bf The performance evaluation.}
We evaluate VGG-16 based on PCNN with sparsities of 55.6\% (n = 4), 66.7\% (n = 3), 77.8\% (n = 2), and 88.9\% (n = 1).
The average activation sparsity is 0.8.
The results show that we can achieve 2.3$\times$, 3.1$\times$, 4.5$\times$, and 9.0$\times$ speedup compared to the dense counterpart. With 256 MAC units running in 300~MHz frequency and 1V voltage, our pattern-aware architecture achieves 3.15~TOPS/W (no sparsity) $\sim$ 28.39~TOPS/W (88.9\% sparsity) power efficiency.
It can be observed that our patter-aware architecture can fully leverage the strengths of our PCNN method.

% \begin{table}[!ht]
%   \centering
%   \setlength{\tabcolsep}{1.1mm}
%   \begin{tabular}{lllll}
%     \hline
%     \textbf{Component} & \textbf{Area ($mm^{2}$)} & \textbf{(\%)} & \textbf{Power($mW$)} & \textbf{(\%)}\\
%     \hline
%     \hline
%     Overall  & 8.00 & 100\% & 99.1 & 100\%\\
%     \hline
%     Data SRAM & 3.25 & 40.6\% & 27.9 & 28.2\%\\
%     % \hline
%     Weight SRAM & 2.48 & 31.0\% & 31.8 & 32.1\%\\
%     % \hline
%     Pattern SRAM & 0.19 & 2.4\% & 1.9 & 1.9\%\\
%     % \hline
%     Register File & 1.58 & 19.8\% & 27.1 & 27.4\%\\
%     % \hline
%     PE group (64 PEs) & 0.50 & 6.2\% & 10.4 & 10.0\%\\
%     \hline
%   \end{tabular}
%   \caption{Area and power characteristics of the chip (not including PLL).}
%   \label{table:layout}
% \end{table}

\begin{table}[!ht]
  \centering
  \setlength{\tabcolsep}{1.1mm}
  \begin{tabular}{lllll}
    \hline
    \textbf{Component} & \textbf{Area ($mm^{2}$)} & \textbf{Share} & \textbf{Power($mW$)} & \textbf{Share}\\
    \hline
    \hline
    Overall  & 8.00 & 100\% & 48.7 & 100\%\\
    \hline
    Data SRAM & 3.25 & 40.6\% & 13.7 & 28.2\%\\
    % \hline
    Weight SRAM & 2.48 & 31.0\% & 15.6 & 32.1\%\\
    % \hline
    Pattern SRAM & 0.19 & 2.4\% & 0.9 & 1.9\%\\
    % \hline
    Register File & 1.58 & 19.8\% & 13.6 & 27.4\%\\
    % \hline
    PE group & 0.50 & 6.2\% & 4.9 & 10.0\%\\
    \hline
  \end{tabular}
  \caption{Area and power characteristics of the chip (not including PLL and IO).}
  \label{table:layout}
  \vspace{-5mm}
\end{table}
\section{Conclusion}

We present PCNN, a novel fine-grained regular pruning method that uses SPM to encode the sparsity.
Contrary to irregular pruning, the sparsity of every kernel is the same in each layer, which can achieve regularity and maintain fine granularity.
Experiments show that 8$\times$ $\sim$ 9$\times$ compression rate and computing speed-up can be achieved with less than 1\% accuracy loss.
Additionally, pattern pruning can be easily combined with coarse-grained pruning methods, achieving 34.4$\times$ compression ratio with negligible accuracy loss.
With SPM, we can deploy indices at the kernel level rather than weight level, which helps to save a great amount of memory overhead.
For computation, with only 3.1\% memory overhead for indices, the proposed architecture can achieve full parallelism and obtain up to 9$\times$ speedup and 28.39~TOPS/W efficiency based on the PCNN.

\bibliographystyle{unsrt}
\bibliography{references}

\begin{thebibliography}{10}

\bibitem{ImageNet}
Alex Krizhevsky, Ilya Sutskever, and Geoffrey~E. Hinton.
\newblock Imagenet classification with deep convolutional neural networks.
\newblock {\em Commun. ACM}, 60(6):84--90, May 2017.

\bibitem{object_detection}
Ross~B. Girshick, Jeff Donahue, Trevor Darrell, and Jitendra Malik.
\newblock Rich feature hierarchies for accurate object detection and semantic
  segmentation.
\newblock {\em CoRR}, abs/1311.2524, 2013.

\bibitem{nlp}
Deng Li Yu Dong Dahl George Mohamed Abdel-rahman \& Jaitly~Navdeep Hinton,
  Geoffrey.
\newblock Deep neural networks for acoustic modeling in speech recognition: The
  shared views of four research groups.
\newblock {\em IEEE Signal Processing Magazine}, 29(6):82--97, Nov 2012.

\bibitem{ResNet}
Kaiming He, Xiangyu Zhang, Shaoqing Ren, and Sun Jian.
\newblock Deep residual learning for image recognition.
\newblock In {\em 2016 IEEE Conference on Computer Vision and Pattern
  Recognition (CVPR)}, 2016.

\bibitem{VGG}
Yoshua Bengio and Yann LeCun, editors.
\newblock {\em 3rd International Conference on Learning Representations, {ICLR}
  2015, San Diego, CA, USA, May 7-9, 2015, Conference Track Proceedings}, 2015.

\bibitem{DianNao}
Tianshi Chen, Zidong Du, Ninghui Sun, Jia Wang, Chengyong Wu, Yunji Chen, and
  Olivier Temam.
\newblock Diannao: A small-footprint high-throughput accelerator for ubiquitous
  machine-learning.
\newblock In {\em Proceedings of the 19th International Conference on
  Architectural Support for Programming Languages and Operating Systems},
  ASPLOS '14, pages 269--284, New York, NY, USA, 2014. ACM.

\bibitem{Eyeriss}
Yu~Hsin Chen, Tushar Krishna, Joel~S. Emer, and Vivienne Sze.
\newblock Eyeriss: An energy-efficient reconfigurable accelerator for deep
  convolutional neural networks.
\newblock {\em IEEE Journal of Solid-State Circuits}, 52(1):127--138, Jan 2017.

\bibitem{UNPU}
Jinmook Lee, Changhyeon Kim, Sanghoon Kang, Dongjoo Shin, Sangyeob Kim, and
  Hoi{-}Jun Yoo.
\newblock {UNPU:} an energy-efficient deep neural network accelerator with
  fully variable weight bit precision.
\newblock {\em J. Solid-State Circuits}, 54(1):173--185, 2019.

\bibitem{FPGA2}
Caiwen Ding, Shuo Wang, Ning Liu, Kaidi Xu, Yanzhi Wang, and Yun Liang.
\newblock {REQ-YOLO:} {A} resource-aware, efficient quantization framework for
  object detection on fpgas.
\newblock In {\em Proceedings of the 2019 {ACM/SIGDA} International Symposium
  on Field-Programmable Gate Arrays, {FPGA} 2019, Seaside, CA, USA, February
  24-26, 2019}, pages 33--42, 2019.

\bibitem{Deep_Compression}
Song Han, Huizi Mao, and William~J. Dally.
\newblock Deep compression: Compressing deep neural network with pruning,
  trained quantization and huffman coding.
\newblock In {\em 4th International Conference on Learning Representations,
  {ICLR} 2016, San Juan, Puerto Rico, May 2-4, 2016, Conference Track
  Proceedings}, 2016.

\bibitem{Prune0}
Suraj Srinivas and R.~Venkatesh Babu.
\newblock Data-free parameter pruning for deep neural networks.
\newblock In {\em Proceedings of the British Machine Vision Conference 2015,
  {BMVC} 2015, Swansea, UK, September 7-10, 2015}, pages 31.1--31.12, 2015.

\bibitem{EIE}
Song Han, Xingyu Liu, Huizi Mao, Jing Pu, Ardavan Pedram, Mark~A. Horowitz, and
  William~J. Dally.
\newblock {EIE:} efficient inference engine on compressed deep neural network.
\newblock In {\em 43rd {ACM/IEEE} Annual International Symposium on Computer
  Architecture, {ISCA} 2016, Seoul, South Korea, June 18-22, 2016}, pages
  243--254, 2016.

\bibitem{prune_regular}
Huizi Mao, Song Han, Jeff Pool, Wenshuo Li, Xingyu Liu, Yu~Wang, and William~J.
  Dally.
\newblock Exploring the granularity of sparsity in convolutional neural
  networks.
\newblock In {\em 2017 {IEEE} Conference on Computer Vision and Pattern
  Recognition Workshops, {CVPR} Workshops 2017, Honolulu, HI, USA, July 21-26,
  2017}, pages 1927--1934, 2017.

\bibitem{regular2}
Sajid Anwar, Kyuyeon Hwang, and Wonyong Sung.
\newblock Structured pruning of deep convolutional neural networks.
\newblock {\em ACM Journal on Emerging Technologies in Computing Systems
  (JETC)}, 13(3):32:1--32:18, 2017.

\bibitem{SSL}
Wei Wen, Chunpeng Wu, Yandan Wang, Yiran Chen, and Hai Li.
\newblock Learning structured sparsity in deep neural networks.
\newblock In {\em Advances in Neural Information Processing Systems 29: Annual
  Conference on Neural Information Processing Systems 2016, December 5-10,
  2016, Barcelona, Spain}, pages 2074--2082, 2016.

\bibitem{CIFAR}
Alex Krizhevsky, Geoffrey Hinton, et~al.
\newblock Learning multiple layers of features from tiny images.
\newblock Technical report, Citeseer, 2009.

\bibitem{boyd2011distributed}
Stephen Boyd, Neal Parikh, Eric Chu, Borja Peleato, Jonathan Eckstein, et~al.
\newblock Distributed optimization and statistical learning via the alternating
  direction method of multipliers.
\newblock {\em Foundations and Trends{\textregistered} in Machine learning},
  3(1):1--122, 2011.

\bibitem{Depth-1}
Hao Li, Asim Kadav, Igor Durdanovic, Hanan Samet, and Hans~Peter Graf.
\newblock Pruning filters for efficient convnets.
\newblock In {\em 5th International Conference on Learning Representations,
  {ICLR} 2017, Toulon, France, April 24-26, 2017, Conference Track
  Proceedings}, 2017.

\bibitem{Depth-2}
Zhuang Liu, Jianguo Li, Zhiqiang Shen, Gao Huang, Shoumeng Yan, and Changshui
  Zhang.
\newblock Learning efficient convolutional networks through network slimming.
\newblock In {\em {IEEE} International Conference on Computer Vision, {ICCV}
  2017, Venice, Italy, October 22-29, 2017}, pages 2755--2763, 2017.

\bibitem{Depth-4}
Qiangui Huang, Shaohua~Kevin Zhou, Suya You, and Ulrich Neumann.
\newblock Learning to prune filters in convolutional neural networks.
\newblock In {\em 2018 {IEEE} Winter Conference on Applications of Computer
  Vision, {WACV} 2018, Lake Tahoe, NV, USA, March 12-15, 2018}, pages 709--718,
  2018.

\bibitem{IKR}
Maurice Yang, Mahmoud Faraj, Assem Hussein, and Vincent~C. Gaudet.
\newblock Efficient hardware realization of convolutional neural networks using
  intra-kernel regular pruning.
\newblock In {\em 48th {IEEE} International Symposium on Multiple-Valued Logic,
  {ISMVL} 2018, Linz, Austria, May 16-18, 2018}, pages 180--185, 2018.

\bibitem{resnet-2}
Adam Dziedzic, John Paparrizos, Sanjay Krishnan, Aaron~J. Elmore, and
  Michael~J. Franklin.
\newblock Band-limited training and inference for convolutional neural
  networks.
\newblock In {\em Proceedings of the 36th International Conference on Machine
  Learning, {ICML} 2019, 9-15 June 2019, Long Beach, California, {USA}}, pages
  1745--1754, 2019.

\bibitem{StructADMM}
Yanzhi Wang, Shaokai Ye, Zhezhi He, Xiaolong Ma, Linfeng Zhang, Sheng Lin, Geng
  Yuan, Sia~Huat Tan, Zhengang Li, Deliang Fan, Xuehai Qian, Xue Lin, and
  Kaisheng Ma.
\newblock Non-structured {DNN} weight pruning considered harmful.
\newblock {\em CoRR}, abs/1907.02124, 2019.

\bibitem{VGG-channel}
Namhoon Lee, Thalaiyasingam Ajanthan, and Philip H.~S. Torr.
\newblock Snip: single-shot network pruning based on connection sensitivity.
\newblock In {\em 7th International Conference on Learning Representations,
  {ICLR} 2019, New Orleans, LA, USA, May 6-9, 2019}, 2019.

\bibitem{Synaptic-Strength}
Chen Lin, Zhao Zhong, Wu~Wei, and Junjie Yan.
\newblock Synaptic strength for convolutional neural network.
\newblock In {\em Advances in Neural Information Processing Systems 31: Annual
  Conference on Neural Information Processing Systems 2018, NeurIPS 2018, 3-8
  December 2018, Montr{\'{e}}al, Canada.}, pages 10170--10179, 2018.

\end{thebibliography}

\end{document}